%% file: main.tex
\documentclass[conference]{IEEEtran}
\IEEEoverridecommandlockouts

\usepackage{cite}
\usepackage{amsmath,amssymb,amsfonts}
\usepackage{algorithmic}
\usepackage{graphicx}
\usepackage{textcomp}
\usepackage{xcolor}
\usepackage{url}
\usepackage{algorithm}
\usepackage{algorithmic}
\usepackage{booktabs}
\usepackage{subcaption}
\usepackage{amsthm}
\usepackage{dsfont}

\newtheorem{theorem}{Theorem}[section]
\newtheorem{proposition}[theorem]{Proposition}

\newtheorem{corollary}[theorem]{Corollary}
\newtheorem{remark}[theorem]{Remark}
\def\BibTeX{{\rm B\kern-.05em{\sc i\kern-.025em b}\kern-.08em
    T\kern-.1667em\lower.7ex\hbox{E}\kern-.125emX}}
\begin{document}

\title{Isolation-based Spherical Ensemble Representations for Tabular Anomaly Detection
}
\author{\IEEEauthorblockN{Yang Cao}
\IEEEauthorblockA{Great Bay University, China \\
Tsinghua University, China \\
charles.cao@ieee.org}
\and
\IEEEauthorblockN{Sikun Yang}
\IEEEauthorblockA{Great Bay University, China \\
sikunyang@gbu.edu.cn}
\and
\IEEEauthorblockN{Hao Tian}
\IEEEauthorblockA{Nanjing University, China \\
htian@smail.nju.edu.cn}
\and
\IEEEauthorblockN{Kai He}
\IEEEauthorblockA{Dongguan University of Technology, China \\
kaihe1214@163.com}
\and
\IEEEauthorblockN{Lianyong Qi}
\IEEEauthorblockA{China University of Petroleum, China \\
lianyongqi@gmail.com}
\and
\IEEEauthorblockN{Ming Liu}
\IEEEauthorblockA{Deakin University, Australia \\
m.liu@deakin.edu.au}
\and
\IEEEauthorblockN{Yujiu Yang}
\IEEEauthorblockA{Tsinghua University, China \\
yang.yujiu@sz.tsinghua.edu.cn}
\and
\IEEEauthorblockN{Hong-Kun Zhang}
\IEEEauthorblockA{Great Bay University, China \\
zhanghk@gbu.edu.cn}
}

\author{
  \IEEEauthorblockN{
    Yang Cao\textsuperscript{1,2},
    Sikun Yang\textsuperscript{1\dag}\thanks{\dag \ Corresponding author.},
    Hao Tian\textsuperscript{3},
    Kai He\textsuperscript{4},
    Lianyong Qi\textsuperscript{5},
    Ming Liu\textsuperscript{6},
    Yujiu Yang\textsuperscript{2},
    Hong-Kun Zhang\textsuperscript{7}
  }
  \IEEEauthorblockA{
    \textsuperscript{1} 
School of Computing and Information Technology, Great Bay University, China, 
    \textsuperscript{2} Tsinghua University, China, \\
    \textsuperscript{3} Nanjing University, China, 
    \textsuperscript{4} Dongguan University of Technology, China,
    \textsuperscript{5} China University of Petroleum, China,\\
    \textsuperscript{6} Deakin University, Australia,  
    \textsuperscript{7} School of Sciences, Great Bay University, China\\ 
    \textbf{Email:} charles.cao@ieee.org, sikunyang@gbu.edu.cn, htian@smail.nju.edu.cn, kaihe1214@163.com, \\ lianyongqi@gmail.com, m.liu@deakin.edu.au, yang.yujiu@sz.tsinghua.edu.cn, zhanghk@gbu.edu.cn
  }
}

\maketitle

\input{content/abstract}


\linespread{0.98}
\input{content/intro}
\input{content/related_work}

\input{content/method}
\input{content/proof}
\input{content/experiment}

\input{content/iseriforest}

\input{content/discuss}
\input{content/conclusion}

\section*{Acknowledgment}
\small Yang Cao is supported by Guangdong Basic and Applied Basic Research Foundation (Grant No.2025A1515110204). Sikun Yang is supported by National Natural Science Foundation of China (NSFC) (Grant No.62476047), the Guangdong Provincial Key Laboratory of Mathematical and Neural Dynamical Systems (Grant No.2024B1212010004), the Guangdong-Dongguan Joint Research Fund (Grant No. 2025A1515140098), and Dongguan Key Laboratory for AI and Dynamical Systems.


\bibliographystyle{IEEEtran}
\bibliography{main}


\end{document}

%% file: content/abstract.tex
\begin{abstract}
  Unsupervised tabular anomaly detection is a critical task with applications spanning offensive language detection, network security, and quality control. Despite extensive research, existing unsupervised anomaly detection methods still face fundamental challenges including conflicting distributional assumptions, computational inefficiency, and difficulty handling different anomaly types. To address these problems, we propose ISER (Isolation-based Spherical Ensemble Representations) that extends existing isolation-based methods by using hypersphere radii as a monotonic transformation of local density characteristics while maintaining linear time and constant space complexity w.r.t. the dataset size. ISER constructs ensemble representations where hypersphere radii encode local sparsity through a monotonic transformation of density: smaller radii correspond to dense regions while larger radii correspond to sparse regions. We introduce a novel similarity-based scoring method that measures pattern consistency by comparing ensemble representations against a theoretical anomaly reference pattern. Additionally, we enhance the performance of Isolation Forest by using ISER and adapting the scoring function to address axis-parallel bias and local anomaly detection limitations. Comprehensive experiments on 20 real-world datasets demonstrate ISER's competitive performance over 12 SOTA methods.

\end{abstract}

%% file: content/intro.tex
\section{Introduction}

Anomaly detection aims to identify data points that deviate significantly from normal patterns, with critical applications in offensive language detection, network security, and quality control~\cite{cao2026tad,tang2024substructure,song2023adops,cao2024automation}. The challenge lies in detecting diverse anomaly types without supervision: global anomalies isolated in sparse regions, local anomalies that deviate within their neighborhoods despite appearing normal globally, and dependency anomalies with unusual feature correlations~\cite{han2022adbench,ye2026comprehensive}. This diversity demands methods that can adapt to different contexts without prior knowledge of anomaly types. 

Existing unsupervised methods face fundamental limitations. Distance-based methods~\cite{angiulli2002fast} assume anomalies are far from neighbors, failing in datasets with varying local densities where normal points in sparse regions are incorrectly flagged. Density-based methods like LOF~\cite{breunig2000lof} address this through local density estimation but require quadratic time complexity for neighborhood queries, limiting scalability to large datasets. Deep learning approaches~\cite{ruff2018deep} can capture complex patterns but demand extensive training and lack interpretability. These trade-offs become particularly acute for tabular data, where heterogeneous feature types, diverse scales, and lack of inherent structure make universal assumptions problematic~\cite{shenkar2022anomaly}. 

Isolation-based methods have shown promise by focusing on the principle that anomalies are few and different, making them easier to isolate through random space partitioning~\cite{liu2008isolation,cao2025anomaly}. However, Isolation Forest's axis-parallel splits primarily reflect global sparsity and struggle with local anomalies~\cite{hariri2019extended}. Isolation-based Nearest Neighbors Ensembles (iNNE)~\cite{bandaragoda2018isolation} uses hypersphere partitioning and nearest radius ratio as anomaly scores but loses discriminative power when regions have similar densities. Isolation Distributional Kernel (IDK)~\cite{ting2020isolation,ting2021isolation} achieves strong performance through kernel method but suffers from computational overhead that limits practical scalability.

We propose ISER (Isolation-based Spherical Ensemble Representations), which addresses these challenges through a spherical isolation mechanism using hypersphere radii as a monotonic transformation of local density. Our insight is that hypersphere radii from spherical partitioning provide a monotonic density proxy. When a hypersphere expands from a random point to its nearest neighbor, the radius reflects local crowding. Dense regions yield small radii while sparse regions yield large radii. This enables local density estimation without explicit neighborhood queries. ISER provides two distinct scoring mechanisms: average-based scoring computes anomaly scores based on density-aware coverage across multiple partitions, while similarity-based scoring directly computes similarity with a predefined reference anomaly pattern. Additionally, our framework can enhance the performance of Isolation Forest.

The contributions of this paper are: 
1) introducing ISER that leverages hypersphere radii directly as local density monotonic transformation through ensemble representations. Unlike iNNE and ADERH that directly derive anomaly scores from partitions, ISER transforms isolation partitions into ensemble representations and performs anomaly assessment in the representation space. 2) proposing a novel similarity-based scoring mechanism that compares ensemble representations against theoretical anomaly patterns, providing robust evaluation complementary to average-based scoring. 3) proposing ISER-IF that extends the proposed ensemble representation framework to significantly improve Isolation Forest performance by addressing its axis-parallel bias and local anomaly detection limitations through modified scoring computation adapted to the transformed feature space. 4) demonstrating ISER's superior performance over 12 competitive baseline methods on 20 real-world datasets, while maintaining the computational efficiency of linear time and constant space complexity w.r.t. the dataset size. The code is available at \url{https://github.com/charles-cao/ISER}. 

%% file: content/related_work.tex
\section{RELATED WORKS}
\textbf{Distance-based} methods assume anomalies are isolated from their neighbors. k-NN~\cite{ramaswamy2000efficient} scores points by their distance to the $k$-th nearest neighbor, but struggles in multi-density datasets where normal points in sparse regions are falsely flagged due to naturally large neighbor distances.

\textbf{Density-based} approaches identify anomalies as points in low-density regions. LOF~\cite{breunig2000lof} computes local outlier factors by comparing a point’s local reachability density with that of its neighbors, enabling detection of local anomalies. Extensions include COF~\cite{tang2002enhancing}, which uses chaining distances for better neighborhood continuity, and LOCI~\cite{papadimitriou2003loci}, which introduces multi-granularity analysis with automatic scale selection. However, these methods suffer from quadratic complexity and sensitivity to neighborhood size.

\textbf{Statistical methods} model normal data distributions and flag low-probability instances. HBOS~\cite{goldstein2012histogram} assumes feature independence and combines histogram bin heights across dimensions, while ECOD~\cite{li2022ecod} uses empirical cumulative distributions to estimate tail probabilities without binning, improving robustness.

\textbf{Isolation-based} methods exploit the fact that anomalies require fewer random splits to isolate, and is widely applied in anomaly detection for texts, streams, time series, etc~\cite{cao2025text,cao2024revisiting,cao2026towards,cao2024detecting}. iForest~\cite{liu2008isolation} uses randomized trees and shorter average path lengths to score anomalies. Variants like SCiForest~\cite{liu2010detecting} and Extended iForest~\cite{hariri2019extended} improve splitting criteria, while iNNE~\cite{bandaragoda2018isolation} uses hyperspheres to capture local density ratios with linear complexity. More recently, IDK~\cite{ting2020isolation,ting2021isolation} has incorporated kernel-based approaches, using data-dependent kernels to measure distribution similarities for improved anomaly detection performance. DIF~\cite{xu2023deep} integrates casually initialized networks with isolation principles. ADERH~\cite{duranianomaly} constructs an ensemble of compact hyperspheres from randomly paired points and refines isolation scores via a ratio-based boundary measure.


\textbf{Deep learning} methods capture complex patterns using neural networks. Autoencoders and VAEs~\cite{kingma2013auto} detect anomalies via reconstruction error, though they may fail when anomalies are reconstructible. DeepSVDD~\cite{ruff2018deep} learns a compact latent representation, and LUNAR~\cite{goodge2022lunar} uses GNNs to model local neighborhood structures. DTE~\cite{livernochediffusion} detects anomalies by directly estimating the posterior distribution over diffusion timesteps for input samples. These methods achieve strong performance but require extensive tuning and lack interpretability.

%% file: content/method.tex
\section{Methodology}\label{method}

\begin{table}[t]
    \centering
    \caption{Key symbols and notations}
    \resizebox{0.35\textwidth}{!}{
    \begin{tabular}{ll}
    \toprule
    $x$ & A data point in input space $\mathbb{R}^d$  \\
    $D$ & Dataset containing $n$ points in $\mathbb{R}^d$ \\
    $d$ & Dimensionality of the feature space \\
    $n$ & Number of points in the dataset \\
    $t$ & Number of different partitionings \\
    $\psi$ & Size of each random subset \\
    $H_i$ & The $i$-th partitioning, $i = 1, \ldots, t$ \\
    $\mathcal{D}_i$ & Random subset of ${D}$ for the $i$-th partitioning \\
    $z$ & A point in $\mathcal{D}_i$ serving as hypersphere center \\
    $\theta[z]$ & Hypersphere centered at point $z$ \\
    $r[z]$ & Radius of hypersphere $\theta[z]$ \\
    $\hat{z}_i(x)$ & Center of the smallest hypersphere in $H_i$ covering $x$ \\
    $\phi_i(x)$ & Representation value for $x$ in $H_i$ \\
    $\Phi(x)$ & Ensemble representation vector of point $x$ \\
    $S_{\text{avg}}$ & Average-based scoring\\
    $S_{\text{sim}}$ & Similarity-based scoring \\
    $\mathds{1}$ & An all-ones vector \\
    $|| \cdot||$ & Distance Function \\
    \bottomrule
    \end{tabular}}
    \label{tab:notation}
\end{table}
In this section, we present ISER (Isolation-based Spherical Ensemble Representations) that combines the efficiency of isolation-based approaches with the local density estimation. Our approach consists of three main components: space partitioning, ensemble representation construction, and two anomaly scoring methods.

\begin{figure}[!htbp]
    \centering
    \includegraphics[width=0.98\linewidth]{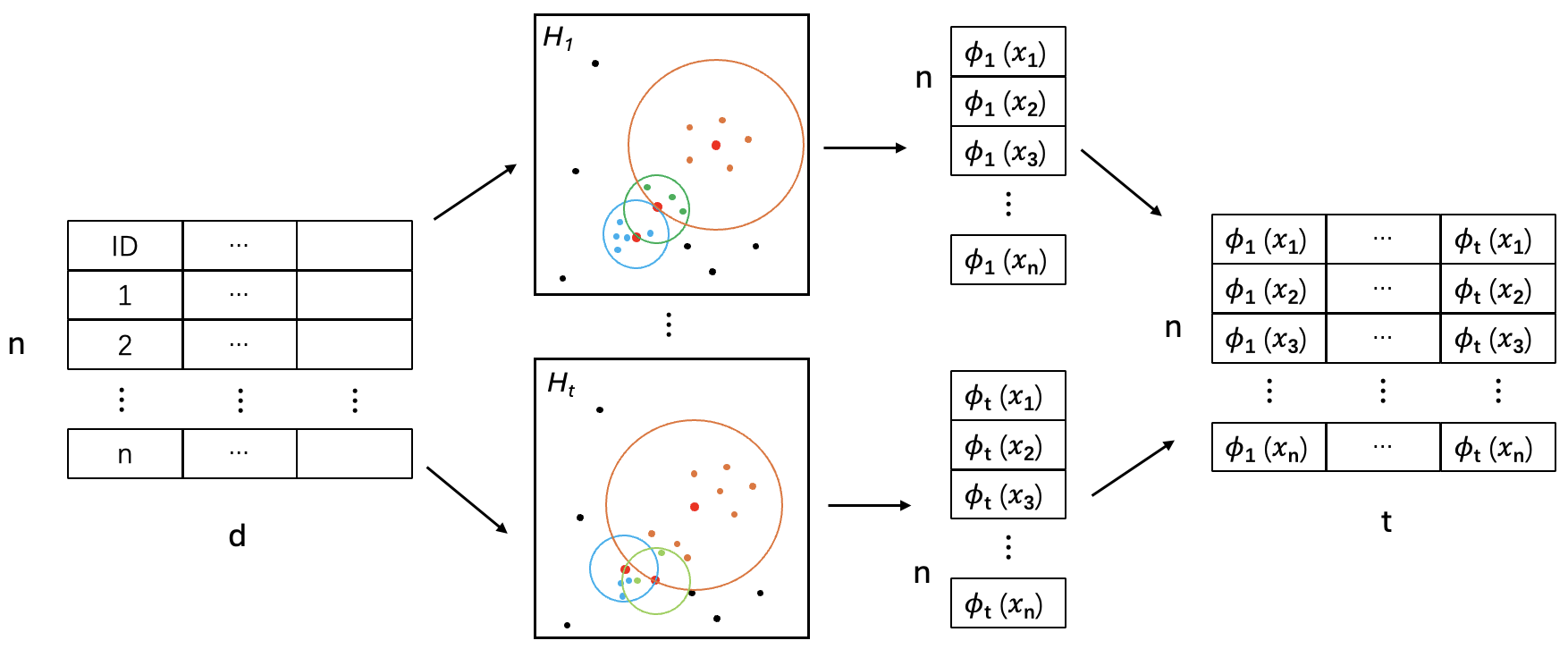}
    \caption{Illustration of ISER ensemble representation construction process. For each partitioning $H_i$, a random subset of $\psi=3$ points is sampled (shown as red bold centers of each circles in the partitions and input data $d=2$ in this sample). Each sampled point serves as a hypersphere center with radius determined by the nearest neighbor distance.  For each point, $\phi_i$ values from all $t$ partitionings are concatenated to form the final ensemble representation $\Phi(x) \in \mathbb{R}^t$.}
    \label{fig:step}
\end{figure}

The core idea of ISER is to partition the feature space using hyperspheres centered at randomly sampled points, where each hypersphere's radius is determined by the distance to each sample point's nearest neighbor within the sampled subset. This partitioning process is repeated multiple times with different random subsets to create an ensemble of independent partitions. In each partitioning, dense regions are more likely to have multiple sampled points, resulting in smaller distances between neighboring sampled points and thus smaller hypersphere radii. Conversely, in sparse regions, fewer points are sampled, leading to larger distances between sampled points and correspondingly larger hypersphere radii. Figure~\ref{fig:step} demonstrates the key steps of ISER, the key notations are shown in Table~\ref{tab:notation}.

\subsection{Space Partitioning and Ensemble Representations}

The proposed ISER employs a hypersphere-based space partitioning mechanism, following the same approach as used in iNNE~\cite{bandaragoda2018isolation} and IDK~\cite{ting2020isolation}.
Given a dataset $D = \{x_1, x_2, \ldots, x_n\} \subset \mathbb{R}^d$, we construct an ensemble of $t$ different partitionings $H_i$, $i = 1, \ldots, t$. For each partitioning $H_i$, we begin by randomly sampling a subset $\mathcal{D}_i \subset {D}$ of size $\psi$ without replacement, $|\mathcal{D}_i| = \psi$ and each partitioning is based on a different random subset.

For each point $z \in \mathcal{D}_i$, we build a hypersphere $\theta[z] \in H_i$ centered at $z$. The radius of this hypersphere is determined by the distance between $z$ and its nearest neighbor in $\mathcal{D}_i \setminus \{z\}$:
\begin{equation}
r[z] = \min_{z' \in \mathcal{D}_i \setminus \{z\}} \|z - z'\|,  \nonumber
\end{equation}

Each partitioning $H_i$ consists of $\psi$ hyperspheres and the region that is not covered by these hyperspheres. Each hypersphere represents a spherical region in the feature space that contains all points within a certain distance from its center. Each hypersphere $\theta[z] \in H_i$ is thus defined by its center $z$ and radius $r[z]$:
\begin{equation}
\theta[z] = \{x \in \mathbb{R}^d : \|x - z\| \leq r[z]\}. \nonumber
\end{equation}

This partitioning mechanism enables the hypersphere radii to serve as a monotonic transformation of local density information. Since the radius of each hypersphere is determined by the nearest neighbor distance within the sampled subset, it directly reflects the local density characteristics of the surrounding region. \textbf{Small radii indicate that sampled points are closely packed, suggesting high local density, while large radii indicate that sampled points are widely separated, suggesting low local density.} 

For any data point $x \in \mathbb{R}^d$, we construct its ensemble representation by evaluating its relationship with each partitioning $H_i$. The key insight is to encode both the coverage information, indicating whether a point falls within any hypersphere, and the density information, captured by the hypersphere radius.


For each partitioning $H_i$ and point $x$, let
$\mathcal{C}_i(x) = \{\, z \in \mathcal{D}_i : \|x - z\| \leq r[z] \,\}$
denote the set of hyperspheres in $H_i$ that cover $x$, we select the covering hypersphere
with the smallest radius:
\begin{equation}
\hat{z}_i(x) = \arg\min_{z \in \mathcal{C}_i(x)} r[z] \nonumber
\end{equation}

For points at boundaries between regions of different densities, a single partitioning assigns the score based on the smallest covering hypersphere. However, the ensemble mechanism provides robustness: across $t$ independent partitionings, boundary points are sometimes assigned to hyperspheres from the dense region (low $\phi$) and sometimes from the sparse region (high $\phi$), with aggregated scores naturally reflecting their transitional position. 

We then compute $\phi_i(x)$, which directly encodes the density characteristics based on hypersphere radius (as opposed to iNNE's local relative measure~\cite{bandaragoda2018isolation}), for point $x$ in $H_i$:

\begin{equation}~\label{e4}
\phi_i(x) =
\begin{cases}
1 - \dfrac{1}{r[\hat{z}_i(x)]} & \text{if } \mathcal{C}_i(x) \neq \emptyset \\
1 & \text{otherwise.} \nonumber
\end{cases}
\end{equation}

The rationale behind this formulation is as follows. When $x$ is covered by at least one hypersphere, we compute $\phi_i(x) = 1 - \frac{1}{r[\hat{z}_i(x)] }$. This transformation preserves density ordering: smaller radii (characteristic of denser regions) yield smaller $\phi$ values, while larger radii (characteristic of sparser regions) yield larger $\phi$ values approaching 1.

When $x$ is covered by no hypersphere in $H_i$, we assign $\phi_i(x) = 1$, indicating that the point is not covered by any hyperspheres in this partitioning. This design choice deliberately ignores local density characteristics for points outside hyperspheres, which is crucial for detecting local anomalies. Through the ensemble of multiple independent partitionings, local anomalies will more frequently fall outside hyperspheres compared to normal points, resulting in consistently higher $\phi$ values regardless of their surrounding density characteristics.

The ensemble representation of point $x$ is constructed by concatenating the coverage indicators from all $t$ partitionings:
\begin{equation}
\Phi(x) = [\phi_1(x), \phi_2(x), \ldots, \phi_t(x)] \in \mathbb{R}^t.  \nonumber
\end{equation}



\subsection{Anomaly Scoring}

\textbf{Average-based Scoring}
The average-based scoring method which we refer to as ISER-A provides a direct approach to anomaly scoring by leveraging the average of the ensemble representation. This approach is grounded in the observation that normal points tend to reside in denser regions and thus consistently exhibit lower ensemble values across multiple partitionings, while anomalous points in sparse regions consistently produce higher values.

The average-based anomaly score is computed as the arithmetic mean of all ensemble components, effectively measuring the average degree of isolation across all partitionings:
\begin{equation}
S_{\text{avg}}(x) = \frac{1}{t}\sum_{i=1}^t \Phi_i(x).  \nonumber
\end{equation}

The interpretation of $S_{\text{avg}}(x)$ is intuitive and directly related to local density characteristics. Higher scores indicate that the point consistently falls outside hyperspheres or within hyperspheres with large radii, both scenarios suggesting the point resides in sparse regions typical of anomalous behavior. Lower scores indicate that the point frequently falls within hyperspheres with small radii, suggesting the point resides in dense regions characteristic of normal behavior. The score is bounded above by 1, where values close to 1 correspond to maximally isolated points and lower values correspond to points in denser regions.

\textbf{Similarity-based Scoring}
Average-based scoring collapses the ensemble representation into a single
scalar by taking the mean. While simple and widely used in isolation-based
ensembles, this aggregation discards two pieces of information: which
partitionings indicate isolation and how consistently they do so. It retains
only the average degree of isolation. As a result, a point that is mildly
isolated across all partitionings and a point that is strongly isolated in
most partitionings but occasionally falls into a dense hypersphere due to
random sampling may receive similar averages, even though only the latter
exhibits the consistent isolation characteristic of a genuine anomaly.

We therefore propose a similarity-based scoring mechanism, referred to as ISER-S, that does not collapse the representation but instead measures how
consistently a point's ensemble representation points toward an anomalous
pattern across all partitionings. This shifts the criterion from how
isolated a point is on average to whether it is consistently isolated, and
as a by-product makes the score robust to occasional values of $\phi = 1$
caused by incomplete spatial coverage in the random sampling process.

The similarity-based approach begins by defining an ideal anomaly reference pattern that represents the theoretical behavior of a perfect anomaly. We define this anomaly reference vector as:
\begin{equation}
\mathds{1} = [1, 1, \ldots, 1] \in \mathbb{R}^t.  \nonumber
\end{equation}
This vector represents the theoretical pattern of a point that lies outside all hyperspheres across all partitionings, indicating maximum and consistent isolation in every dimension of the ensemble space. Such a pattern would be exhibited by a point that is consistently far from all local density centers, representing the most extreme form of anomalous behavior.

The similarity-based anomaly score is computed using cosine similarity between the point's ensemble representation and the anomaly reference vector:
\begin{equation}
S_{\text{sim}}(x) = \text{sim}(\Phi(x), \mathds{1}).  \nonumber
\end{equation}

\subsection{Scalability Analysis}

Algorithm~\ref{alg:iser} is the pseudo code for ISER.

\textbf{Time complexity:} In the training phase, for each partitioning $H_i$ (with $\psi$ sampled points), nearest neighbor searches are required to compute radii for $\psi$ hyperspheres, each centered at a sampled point. This is done $t$ times to form an ensemble of $t$ partitionings of $\psi$ hyperspheres. The training time complexity is $O(t\psi^2)$. For evaluating a point $x$, we compute its ensemble representation by identifying the smallest covering hypersphere in each partitioning (requiring $O(\psi)$ time per partitioning) and evaluating the coverage indicator. The anomaly score computation takes $O(t)$ time. Therefore, the total inference complexity is $O(n \psi t)$. The overall time complexity is \textbf{linear} with respect to $n$ because $t$ and $\psi$ are constant hyperparameters and $t \psi \ll n$ for large datasets.

\textbf{Space Complexity:} In the training phase, ISER stores the $t\psi$ hypersphere centers and their radii, requiring $O(t\psi d)$ space, which is constant with respect to $n$. In inference phase, ISER computes a $t$-dimensional feature vector to achieve anomaly score for each point. Therefore, the total space complexity is linear with respect to $n$.

\begin{algorithm}[!htbp]
\caption{ISER}
\label{alg:iser}
\begin{algorithmic}[1]
\REQUIRE Dataset $D = \{x_1, x_2, \ldots, x_n\} \subset \mathbb{R}^d$, number of partitionings $t$, sample size $\psi$
\ENSURE Ensemble representation $\Phi(x) \in \mathbb{R}^t$ for any point $x$
\FOR{$i = 1$ to $t$}
    \STATE Randomly sample $\psi$ points from $\mathcal{D}$ without replacement
    \STATE $\mathcal{D}_i = \{z_1, z_2, \ldots, z_\psi\} \subset \mathcal{D}$
    \FOR{each $z \in \mathcal{D}_i$}
        \STATE Compute radius: $r[z] = \min_{z' \in \mathcal{D}_i \setminus \{z\}} \|z - z'\|$
    \ENDFOR
\ENDFOR
\FOR{any data point $x \in \mathbb{R}^d$}
    \FOR{$i = 1$ to $t$}
        \STATE $\mathcal{C}_i(x) \leftarrow \{z \in \mathcal{D}_i : \|x-z\| \leq r[z]\}$
        \IF{$\mathcal{C}_i(x) \neq \emptyset$}
            \STATE $\hat{z}_i(x) \leftarrow \arg\min_{z \in \mathcal{C}_i(x)} r[z]$
            \STATE $\phi_i(x) \leftarrow 1 - \frac{1}{r[\hat{z}_i(x)]}$
        \ELSE
            \STATE $\phi_i(x) \leftarrow 1$
        \ENDIF
    \ENDFOR
    \STATE $\Phi(x) = [\phi_1(x), \phi_2(x), \ldots, \phi_t(x)]$
\ENDFOR
\STATE \textbf{Average-based:} $S_{\text{avg}}(x) = \frac{1}{t}\sum_{i=1}^t \Phi_i(x)$
\STATE \textbf{Similarity-based:} $S_{\text{sim}}(x) = \text{sim}(\Phi(x), \mathds{1})$
\end{algorithmic}
\end{algorithm}


%% file: content/proof.tex

\section{Theoretical Analysis}\label{appendix:theory}
In this section, we provide a theoretical justification for using
hypersphere radii as a monotonic transformation of local density. This
analysis establishes the relationship between the magnitude of $\phi$ and
local density, which directly grounds average-based scoring (ISER-A);
similarity-based scoring (ISER-S) builds on the same representation but
additionally exploits the consistency of the isolation pattern across
partitionings, as discussed in Section VII-A. 

For the theoretical analysis, we define the transformation function 
$f(r) = 1 - 1/r$. From Equation~\ref{e4}, when point $x$ falls within a hypersphere, $\phi_i(x) = 1 - \frac{1}{r[\hat{z}_i(x)]} = f(r[\hat{z}_i(x)])$. We prove that the expected radius of hyperspheres constructed through our random sampling procedure exhibits a monotonic relationship with the underlying data density, thereby validating the use of $f(r) = 1 - 1/r$ for density-aware anomaly detection. 


\subsection{Problem Setup}

Consider a dataset $\mathcal{D} = \{x_1, x_2, \ldots, x_N\} \subset \mathbb{R}^d$ consisting of $N$ data points in a $d$-dimensional feature space. The dataset exhibits spatially varying density characteristics across different regions. For a region $V$ with volume $|V|$ containing $N_V$ data points, we define the local density as $\rho = \frac{N_V}{|V|}$.

Our algorithm constructs hypersphere-based partitions by randomly sampling $\psi$ points from $\mathcal{D}$ (without replacement) to serve as hypersphere centers. Each center $z$ determines a hypersphere radius $r[z]$ based on the distance to its nearest neighbor among the sampled points.

\subsection{Main Theoretical Results}

\begin{proposition}[Sampling Distribution]
\label{prop:sampling}
When randomly sampling $\psi$ points from dataset $\mathcal{D}$, the expected number of sampled points falling in region $V$ is:
\begin{equation}
\mathbb{E}[\psi_V] = \psi \cdot \frac{N_V}{N} = \psi \cdot \frac{\rho |V|}{N}, \nonumber
\end{equation}
where $\psi_V$ denotes the random variable representing the number of sampled points in $V$.
\end{proposition}

\begin{proof}
The sampling process follows a hypergeometric distribution. Each point in $V$ has probability $\psi/N$ of being selected (for large $N$). Therefore:
\begin{equation}
\mathbb{E}[\psi_V] = N_V \cdot \frac{\psi}{N} = \psi \cdot \frac{N_V}{N}. \nonumber
\end{equation}
Substituting $N_V = \rho |V|$ yields the result.
\end{proof}

\begin{corollary}[Sampling Density Preservation]
\label{cor:density_preservation}
The sampling density in region $V$, defined as $\rho^{\text{sample}} = \frac{\psi_V}{|V|}$, satisfies:
\begin{equation}
\mathbb{E}[\rho^{\text{sample}}] = \frac{\psi \rho}{N}. \nonumber
\end{equation}
Consequently, for two regions with densities $\rho_1$ and $\rho_2$:
\begin{equation}
\frac{\mathbb{E}[\rho_1^{\text{sample}}]}{\mathbb{E}[\rho_2^{\text{sample}}]} = \frac{\rho_1}{\rho_2}.  \nonumber
\end{equation}
\end{corollary}

\begin{proof}
By definition, the sampling density in region $V$ is:
\begin{equation}
\rho^{\text{sample}} = \frac{\psi_V}{|V|}.  \nonumber
\end{equation}

Taking expectation on both sides, and noting that $|V|$ is a constant:
\begin{equation}
\mathbb{E}[\rho^{\text{sample}}] = \mathbb{E}\left[\frac{\psi_V}{|V|}\right] = \frac{\mathbb{E}[\psi_V]}{|V|}. \nonumber
\end{equation}

From Proposition~\ref{prop:sampling}, we have $\mathbb{E}[\psi_V] = \psi \cdot N_V / N$. Substituting:
\begin{equation}
\mathbb{E}[\rho^{\text{sample}}] = \frac{\psi N_V / N}{|V|} = \frac{\psi N_V}{N |V|}.  \nonumber
\end{equation}

Since the local density is defined as $\rho = N_V / |V|$, we have $N_V = \rho |V|$. Therefore:
\begin{equation}
\mathbb{E}[\rho^{\text{sample}}] = \frac{\psi \cdot \rho |V|}{N |V|} = \frac{\psi \rho}{N}. \nonumber
\end{equation}

For the density ratio, applying the above result to two regions:
\begin{equation}
\frac{\mathbb{E}[\rho_1^{\text{sample}}]}{\mathbb{E}[\rho_2^{\text{sample}}]} = \frac{\psi \rho_1 / N}{\psi \rho_2 / N} = \frac{\rho_1}{\rho_2}, \nonumber
\end{equation}
where the constants $\psi$ and $N$ cancel out, establishing that sampling preserves relative density ratios.
\end{proof}

\begin{proposition}[Radius-Density Relationship]
\label{prop:radius_density}
For a hypersphere center $z$ in region $V$ with local density $\rho$, the expected nearest neighbor distance among sampled points satisfies:
\begin{equation}
\mathbb{E}[r] \approx C \cdot \rho^{-1/d}, \nonumber
\end{equation}
where $C = K_d^{-1/d} \cdot (\psi/N)^{-1/d}$ is a constant, and $K_d = \pi^{d/2}/\Gamma(d/2+1)$ is the volume constant for $d$-dimensional hyperspheres.
\end{proposition}

\begin{proof}
Consider a sampled point $z$ in region $V$. We model the nearest neighbor distance by analyzing the expected number of other sampled points within a hypersphere of radius $r$ centered at $z$.

\textbf{Step 1: Hypersphere volume.} 
The volume of a $d$-dimensional hypersphere with radius $r$ is given by the standard geometric formula:
\begin{equation}
V_d(r) = K_d \cdot r^d, \quad K_d = \frac{\pi^{d/2}}{\Gamma(d/2+1)}, \nonumber
\end{equation}
where $\Gamma(\cdot)$ denotes the Gamma function, a generalization of the factorial function.
This reduces to familiar formulas: $V_2(r) = \pi r^2$ for circles and $V_3(r) = \frac{4\pi}{3}r^3$ for spheres.

\textbf{Step 2: Expected point count.}
The expected number of other sampled points within the hypersphere is the product of sampling density and volume:
\begin{equation}
\mathbb{E}[\text{\# points in } V_d(r)] = \rho^{\text{sample}} \cdot K_d r^d. \nonumber
\end{equation}

\textbf{Step 3: Nearest neighbor condition.}
The nearest neighbor distance $r$ is defined as the radius at which we expect exactly one other sampled point within the hypersphere:
\begin{equation}
\rho^{\text{sample}} \cdot K_d r^d = 1. \nonumber
\end{equation}

\textbf{Step 4: Solving for radius.}
Rearranging:
\begin{equation}
r^d = \frac{1}{\rho^{\text{sample}} \cdot K_d}. \nonumber
\end{equation}
Taking the $d$-th root:
\begin{equation}
r = \left(\rho^{\text{sample}} \cdot K_d\right)^{-1/d} = K_d^{-1/d} \cdot \left(\rho^{\text{sample}}\right)^{-1/d}. \nonumber
\end{equation}


\textbf{Step 5: Approximation via concentration.}
From Corollary~\ref{cor:density_preservation}, $\mathbb{E}[\rho^{\text{sample}}] = \psi\rho/N$. 
From Step 4, we have $r = K_d^{-1/d} \cdot (\rho^{\text{sample}})^{-1/d}$. 
Taking expectation:
\begin{equation}
\mathbb{E}[r] = \mathbb{E}\left[K_d^{-1/d} \cdot (\rho^{\text{sample}})^{-1/d}\right] 
= K_d^{-1/d} \cdot \mathbb{E}\left[(\rho^{\text{sample}})^{-1/d}\right]. \nonumber
\end{equation}
For moderate to large $\psi$, the random variable $\rho^{\text{sample}}$ concentrates 
tightly around its mean. Under this concentration:
\begin{equation}
\mathbb{E}\left[(\rho^{\text{sample}})^{-1/d}\right] \approx 
\left(\mathbb{E}[\rho^{\text{sample}}]\right)^{-1/d} 
= \left(\frac{\psi\rho}{N}\right)^{-1/d}. \nonumber
\end{equation}
Therefore:
\begin{equation}
\mathbb{E}[r] \approx K_d^{-1/d} \cdot \left(\frac{\psi}{N}\right)^{-1/d} \cdot \rho^{-1/d} 
= C \cdot \rho^{-1/d}, \nonumber
\end{equation}
where $C = K_d^{-1/d} \cdot (\psi/N)^{-1/d}$.
\end{proof}

\begin{corollary}[Monotonic Relationship]
\label{cor:monotonicity}
The transformation $f(r) = 1 - 1/r$ preserves density ordering. Specifically, for two regions with densities $\rho_1 > \rho_2$:
\begin{equation}
\mathbb{E}[f(r_1)] < \mathbb{E}[f(r_2)], \nonumber
\end{equation}
where $r_1$ and $r_2$ are the corresponding hypersphere radii.
\end{corollary}

\begin{proof}
From Proposition~\ref{prop:radius_density}, $\rho_1 > \rho_2$ implies:
\begin{equation}
\mathbb{E}[r_1] \approx C \cdot \rho_1^{-1/d} < C \cdot \rho_2^{-1/d} \approx \mathbb{E}[r_2]. \nonumber
\end{equation}

Given that $\psi$ is moderate to large, both $r_1$ and $r_2$ concentrate around 
their respective expectations. For strictly increasing $f(r) = 1 - 1/r$ and 
concentrated distributions, we have:
\begin{equation}
\mathbb{E}[f(r_1)] \approx f(\mathbb{E}[r_1]) < f(\mathbb{E}[r_2]) \approx \mathbb{E}[f(r_2)], \nonumber
\end{equation}
where the approximations $\mathbb{E}[f(r)] \approx f(\mathbb{E}[r])$ hold due to 
the small variance of $r$ under concentration.

Thus, the transformation preserves the inverse ordering of densities.
\end{proof}

\subsection{Implications for Anomaly Detection}


\begin{remark}[Nonlinearity and Sufficiency]
The dependence of the expected radius on density is nonlinear: $\mathbb{E}[r] \propto \rho^{-1/d}$. 
The scoring function $f(r) = 1 - 1/r$ does not linearize this relationship. 
Nevertheless, density-based anomaly detection methods require preserving the ordering of densities. 
By Corollary~\ref{cor:monotonicity}, $f(r)$ is strictly decreasing in $\rho$, which ensures correct ranking of normal versus anomalous instances. 
Since AUC-based metrics depend solely on score orderings, this monotonicity is sufficient.
\end{remark}

\begin{remark}[Multi-Density Datasets]
Real-world datasets typically contain multiple regions with distinct density characteristics. Our theoretical framework naturally handles such scenarios. Consider a dataset with $m$ regions having densities $\rho_1, \rho_2, \ldots, \rho_m$. Proposition~\ref{prop:radius_density} guarantees that for any pair of regions $i$ and $j$:
\begin{equation}
\rho_i > \rho_j \implies \mathbb{E}[r_i] < \mathbb{E}[r_j] \implies \mathbb{E}[f(r_i)] < \mathbb{E}[f(r_j)]. \nonumber
\end{equation}
When combined with ensemble aggregation over multiple independent partitionings, this pairwise ordering property extends to the entire density spectrum, ensuring that sparser (more anomalous) regions consistently receive higher anomaly scores regardless of the absolute density values.
\end{remark}

\begin{remark}[Boundary Points and Ensemble Aggregation]
For data points located at boundaries between regions of different densities, a single partitioning may assign scores based on the smallest covering hypersphere, which could belong to either the dense or sparse region. However, across the ensemble of $t$ independent random partitionings, boundary points exhibit intermediate behavior:
\begin{itemize}
    \item Some partitionings sample centers predominantly in the dense region, assigning low anomaly scores.
    \item Other partitionings sample centers in the sparse region, assigning high anomaly scores.
    \item The ensemble aggregation (via averaging or similarity scoring) produces intermediate anomaly scores that appropriately reflect the transitional nature of boundary points.
\end{itemize}
This stochastic averaging mechanism provides robustness to local variations in density and reduces sensitivity to individual partitioning artifacts.
\end{remark}

\subsection{Coverage-Based Encoding for Local Anomalies}

While Proposition~\ref{prop:radius_density} addresses radius-based encoding for points within hyperspheres, our method assigns $\phi_i(x) = 1$ when point $x$ falls outside all hyperspheres in partitioning $H_i$. This coverage-based encoding is particularly effective for detecting local anomalies.

\begin{remark}[Local Anomaly Detection Mechanism]
Local anomalies are characterized by deviating from their local neighborhood structure while not necessarily being globally isolated. Consider a local anomaly $x^*$ near a dense region with high density $\rho_{\text{dense}}$. Through random sampling:
\begin{enumerate}
    \item By Corollary~\ref{cor:density_preservation}, the dense region receives a high concentration of sampled centers.
    \item These centers form hyperspheres that cover the normal data manifold within the dense region.
    \item The local anomaly $x^*$, while spatially proximate, lies off this manifold and is less likely to be covered by hyperspheres centered on normal points.
    \item Across $t$ independent partitionings, $x^*$ exhibits higher frequency of non-coverage ($\phi = 1$) compared to normal points in the same spatial vicinity.
\end{enumerate}
This mechanism allows ISER to distinguish local anomalies from nearby normal points, addressing a key limitation of global density-based methods.
\end{remark}

In summary, the theoretical analysis establishes that:
\begin{enumerate}
    \item Random sampling induces a sampling density that preserves the relative ordering of local densities (Proposition~\ref{prop:sampling}, Corollary~\ref{cor:density_preservation})
    \item Hypersphere radii exhibit a power-law relationship with local density: $\mathbb{E}[r] \propto \rho^{-1/d}$ (Proposition~\ref{prop:radius_density})
    \item The transformation $f(r) = 1 - 1/r$ maintains density ordering despite being nonlinear (Corollary~\ref{cor:monotonicity})
    \item This monotonic relationship is sufficient for anomaly detection as a ranking task, where relative ordering rather than absolute density values determines performance
\end{enumerate}

%% file: content/experiment.tex
\section{Experiment}

\subsection{Experiment Setup}
\begin{table}[!htbp]
\centering
\caption{Statistical information of datasets.}
\label{tab:data}
\resizebox{0.3\textwidth}{!}{
\begin{tabular}{lllll}
\toprule
Dataset      & \#Samples   & \#Features & \#Ano. & \%Ano.  \\
\midrule
Annthyroid   & 7200   & 6    & 534   & 7.42   \\
backdoor     & 95329  & 196  & 2329  & 2.44   \\
breastw      & 683    & 9    & 239   & 34.99  \\
celeba       & 202599 & 39   & 4547  & 2.24   \\
fault        & 1941   & 500  & 673 & 34.67 \\
glass        & 214    & 7    & 9     & 4.21   \\
Ionosphere   & 351    & 32   & 126   & 35.90  \\
Landsat      & 6435   & 36   & 1333  & 20.71  \\
Lymphography & 148    & 18   & 6     & 4.05   \\
mnist        & 7603   & 100  & 700   & 9.21   \\
musk         & 3062   & 166  & 97    & 3.17   \\
pendigits    & 6870   & 16   & 156   & 2.27   \\
Pima         & 768    & 8    & 268   & 34.90  \\
satellite    & 6435   & 36   & 2036  & 31.64  \\
satimage-2   & 5803   & 36   & 71    & 1.22   \\
shuttle      & 49094  & 9    & 3511  & 7.15   \\
skin         & 245057 & 3    & 50859 & 20.75  \\
speech       & 3686   & 400  & 61    & 1.65   \\
vowels       & 1456   & 12   & 50    & 3.43   \\
WPBC         & 198    & 33   & 47    & 23.74 \\
\bottomrule
\end{tabular}}
\end{table}

\textbf{Problem Setting.} 
Following the unsupervised setting, our experiments assume access to the entire dataset containing both normal and anomalous instances during training, without requiring a clean training set of only normal samples. This differs from one-class settings where methods assume access to pure normal data during training. All methods are evaluated on the same contaminated datasets to ensure fair comparison.



\textbf{Datasets.} 
We conduct experiments on 20 real-world benchmark datasets from PyOD~\cite{zhao2019pyod} and ODDS~\cite{Rayana:2016} repositories. 
The datasets exhibit diverse characteristics in terms of size, dimensionality and contamination rates, as detailed in Table~\ref{tab:data}. They are from different domains including healthcare, geology, astronautics, etc.

\textbf{Baseline Methods.} We compare ISER against 12 representative methods: LOF~\cite{breunig2000lof}, iForest~\cite{liu2008isolation}, ECOD~\cite{li2022ecod}, COPOD~\cite{li2020copod}, DeepSVDD~\cite{ruff2018deep}, AutoEncoder, LUNAR~\cite{goodge2022lunar}, DIF~\cite{xu2023deep}, DTE-C~\cite{livernochediffusion}, iNNE~\cite{bandaragoda2018isolation}, IDK~\cite{ting2021isolation} and ADERH~\cite{duranianomaly}. We evaluate both  ISER-A and ISER-S.

\textbf{Hyperparameter settings.} The hyperparameter of nearest neighbors for LOF and LUNAR is searched in $\{5, 10, 20, 40\}$. For iForest, iNNE, IDK, ADERH and proposed methods, $\psi \in \{2^m, m=1,2,\dots,8\}$ and with default $t=200$. For Autoencoder, DIF and DeepSVDD, the hyperparameter of hidden neuron is searched in $\{[128, 64], [64, 32], [32, 16]\}$. ECOD and COPOD have no parameters.

\textbf{Evaluation Metrics.} We use Area Under the ROC Curve (AUC-ROC) as evaluation metrics. Each experiment is repeated 5 times and reported average results.

\subsection{Empirical Evaluation on Real-World Datasets}
\begin{table*}[!htbp]
\centering
\caption{Results of all methods in terms of AUC-ROC. The best average performance is shown in bold.}
\label{tab:AUC-ROC}
\resizebox{0.8\textwidth}{!}{%
\begin{tabular}{lllllllllllll|ll}
\hline
\textbf{}             & \textbf{LOF} & \textbf{iForest} & \textbf{ECOD} & \textbf{COPOD} & \textbf{D.SVDD} & \textbf{AE} & \textbf{LUNAR} & \textbf{DIF} & \textbf{DTE-C} & \textbf{INNE} & \textbf{IDK} & \textbf{ADERH} & \textbf{ISER-A} & \textbf{ISER-S} \\ \hline
\textbf{Annthyroid}   & 0.7362       & 0.8608           & 0.7887        & 0.7760         & 0.6964            & 0.7798      & 0.7367         & 0.6758       & 0.9636       & 0.7455        & 0.7231       & 0.7667         & 0.8270          & 0.8030          \\
\textbf{backdoor}     & 0.7782       & 0.7516           & 0.8462        & 0.7893         & 0.9026            & 0.9164      & 0.6036         & 0.9189       & 0.8752       & 0.9044        & 0.9044       & 0.8986         & 0.9120          & 0.9162          \\
\textbf{breastw}      & 0.4691       & 0.9953           & 0.9914        & 0.9944         & 0.9338            & 0.9742      & 0.9780         & 0.7989       & 0.8907       & 0.9822        & 0.9941       & 0.9904         & 0.9853          & 0.9859          \\
\textbf{celeba}       & 0.4975       & 0.7049           & 0.7572        & 0.7508         & 0.7350            & 0.7374      & 0.5771         & 0.4993       & 0.8122       & 0.8059        & 0.8315       & 0.8155         & 0.8088          & 0.8376          \\
\textbf{fault}        & 0.5973       & 0.5912           & 0.4687        & 0.4553         & 0.5048            & 0.6898      & 0.7253         & 0.7004       & 0.5895       & 0.6174        & 0.7023       & 0.7302         & 0.7210          & 0.7314          \\
\textbf{glass}        & 0.8022       & 0.8016           & 0.7046        & 0.7550         & 0.6759            & 0.8439      & 0.8525         & 0.8546       & 0.8639       & 0.7954        & 0.8046       & 0.8643         & 0.8730          & 0.8805          \\
\textbf{Ionosphere}   & 0.8934       & 0.8530           & 0.7284        & 0.7895         & 0.7850            & 0.9459      & 0.9281         & 0.8975       & 0.9114       & 0.8899        & 0.8883       & 0.9247         & 0.9179          & 0.9359          \\
\textbf{Landsat}      & 0.5466       & 0.4945           & 0.3678        & 0.4215         & 0.3960            & 0.5168      & 0.6070         & 0.5650       & 0.5442       & 0.6139        & 0.6988       & 0.6158         & 0.5963          & 0.6316          \\
\textbf{Lymphography} & 0.9953       & 0.9995           & 0.9953        & 0.9965         & 0.9932            & 0.9930      & 0.9899         & 0.9453       & 0.8341       & 0.9967        & 0.9981       & 0.9960         & 0.9969          & 0.9965          \\
\textbf{mnist}        & 0.7380       & 0.8132           & 0.7463        & 0.7739         & 0.8428            & 0.8570      & 0.8079         & 0.8825       & 0.8189       & 0.8627        & 0.8708       & 0.8687         & 0.8688          & 0.8849          \\
\textbf{musk}         & 0.6357       & 1.0000           & 0.9559        & 0.9463         & 0.9727            & 0.9780      & 0.8543         & 0.9894       & 0.9647       & 1.0000        & 1.0000       & 1.0000         & 1.0000          & 1.0000          \\
\textbf{pendigits}    & 0.5240       & 0.9517           & 0.9274        & 0.9048         & 0.8581            & 0.8661      & 0.7139         & 0.9560       & 0.7133       & 0.9140        & 0.9340       & 0.9576         & 0.9358          & 0.9619          \\
\textbf{Pima}         & 0.6399       & 0.6934           & 0.5944        & 0.6540         & 0.6781            & 0.6809      & 0.7053         & 0.6306       & 0.6237       & 0.7010        & 0.7306       & 0.7343         & 0.7118          & 0.7356          \\
\textbf{satellite}    & 0.5601       & 0.7149           & 0.5830        & 0.6335         & 0.6049            & 0.6847      & 0.6672         & 0.7405       & 0.7105       & 0.7345        & 0.7727       & 0.7994         & 0.7658          & 0.7861          \\
\textbf{satimage-2}   & 0.5885       & 0.9945           & 0.9649        & 0.9745         & 0.9619            & 0.9946      & 0.8819         & 0.9971       & 0.9456       & 0.9976        & 0.9983       & 0.9989         & 0.9981          & 0.9987          \\
\textbf{shuttle}      & 0.5289       & 0.9972           & 0.9929        & 0.9945         & 0.9879            & 0.9948      & 0.6752         & 0.9751       & 0.9762       & 0.9923        & 0.9916       & 0.9936         & 0.9928          & 0.9934          \\
\textbf{skin}         & 0.5532       & 0.6808           & 0.4888        & 0.4711         & 0.5603            & 0.7054      & 0.7157         & 0.6598       & 0.7405       & 0.7492        & 0.7833       & 0.7877         & 0.7779          & 0.7889          \\
\textbf{speech}       & 0.6520       & 0.4803           & 0.4697        & 0.4911         & 0.5286            & 0.4760      & 0.5114         & 0.4934       & 0.4946       & 0.6462        & 0.6892       & 0.5095         & 0.6437          & 0.6952          \\
\textbf{vowels}       & 0.9434       & 0.7647           & 0.5929        & 0.4958         & 0.5884            & 0.9378      & 0.9336         & 0.8268       & 0.9142       & 0.9415        & 0.9528       & 0.9811         & 0.9468          & 0.9588          \\
\textbf{WPBC}         & 0.5201       & 0.5215           & 0.4813        & 0.5233         & 0.5159            & 0.5125      & 0.5204         & 0.4743       & 0.4679       & 0.5080        & 0.5201       & 0.5273         & 0.5173          & 0.5240          \\ \hline
\textbf{Avg.}         & 0.6600       & 0.7832           & 0.7223        & 0.7296         & 0.7361            & 0.8043      & 0.7493         & 0.7741       & 0.7827       & 0.8199        & 0.8394       & 0.8380         & 0.8399          & \textbf{0.8523}     \\ \hline    
\end{tabular}%
}
\end{table*}

Table~\ref{tab:AUC-ROC} presents comprehensive performance comparisons across 20 real-world datasets. The results demonstrate that ISER methods consistently achieve competitive performance across diverse data characteristics. ISER-S achieves the highest average AUC-ROC of 0.8523, outperforming all baseline methods including the most competitive isolation-based method IDK. ISER-A also demonstrates strong performance with an average AUC-ROC of 0.8399, positioning both ISER variants among the top-performing methods. Notably, ISER-S exhibits robust performance across datasets with varying characteristics. On high-dimensional datasets such as fault and speech, ISER methods maintain competitive detection capability. For large-scale datasets such as backdoor, celeba and skin, ISER-S achieves better detection performance comparable to deep learning approaches while maintaining computational efficiency.

\begin{figure}[!htbp]
     \centering
     \includegraphics[width=0.33\textwidth]{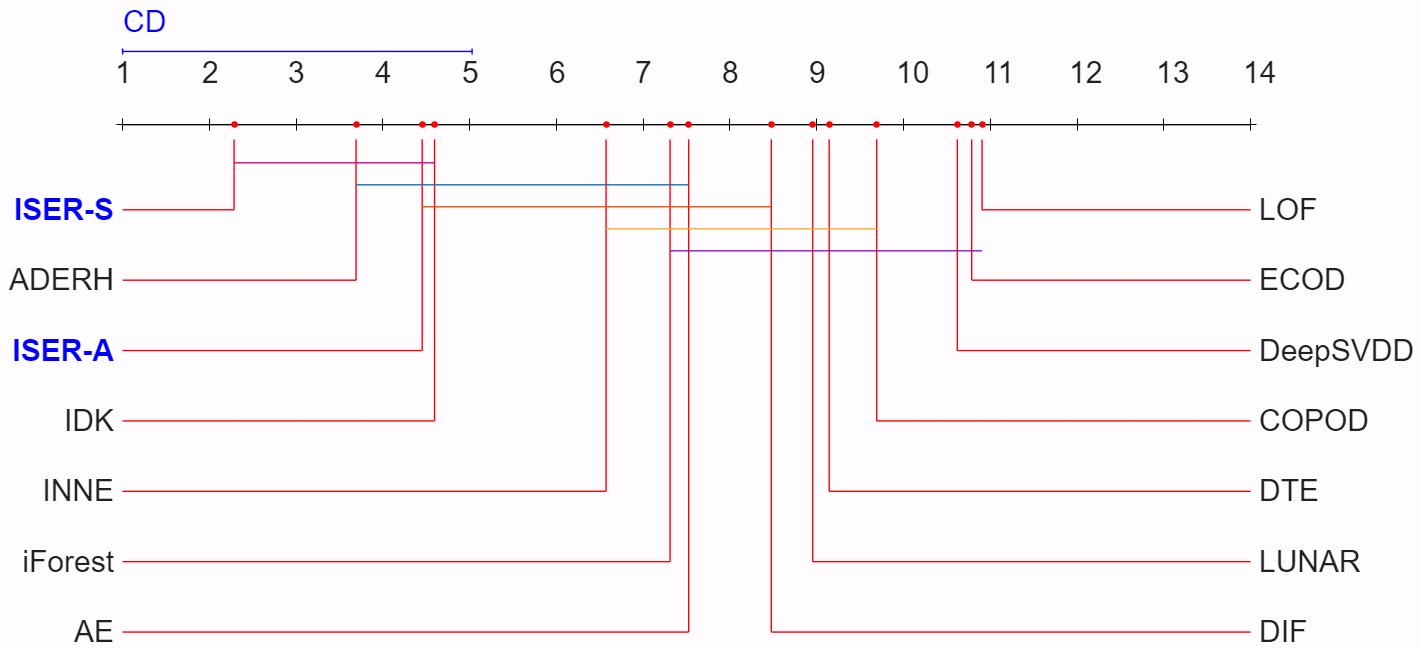} 
     \caption{Friedman-Nemenyi test for anomaly detectors at significance level $0.1$. If two algorithms are connected by a CD (critical difference) line, there is no significant difference between them.}
     \label{sig}
\end{figure}

The critical difference diagram in Figure~\ref{sig} provides Friedman-Nemenyi test at significance level $0.1$. ISER-S is the only method that shows statistical significance over all methods ranked below IDK. 

\subsection{Detection of Local Anomalies}

\begin{figure}[!htbp]
     \centering
     \begin{subfigure}[b]{0.2\textwidth}
         \centering
         \includegraphics[width=0.6\textwidth]{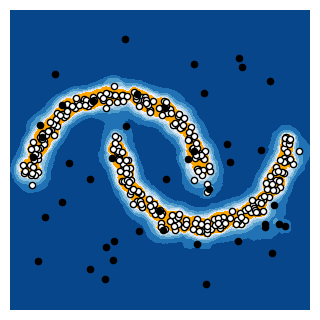}
         \caption{ISER-A $= 0.85$}
         \label{a-local}
     \end{subfigure}
     \begin{subfigure}[b]{0.2\textwidth}
         \centering
         \includegraphics[width=0.6\textwidth]{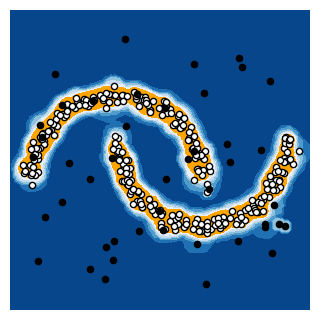}
         \caption{ISER-S $= 0.91$}
         \label{s-local}
         \end{subfigure}
     \caption{Results of ISER on a synthetic two-moons dataset, the orange region is the normal region, white points are normal and black points are anomalies. }
     \label{fig:types}
\end{figure}

To show that ISER detects \emph{local} anomalies, i.e., objects that are
anomalous relative to their local neighborhood rather than only globally
extreme points, we visualize its results on a synthetic two-moons dataset
of 300 points with a 0.15 anomaly rate, where the normal objects form two
interleaving non-convex crescents and the anomalies are scattered across
the space. We flag the 15\% of points with the highest anomaly scores as
anomalies; the orange region in Fig.~\ref{fig:types} shows the resulting
normal region. Because ISER uses the hypersphere radius as a proxy for
local density, the orange region follows the two crescents closely, and
the scattered anomalies are correctly separated from the normal manifold.


\subsection{Sensitivity Analysis}\label{sense}
We analyze ISER's sensitivity to its two key parameters: sample size $\psi$ and number of partitionings $t$.

\begin{figure}[ht]
     \centering
     \begin{subfigure}[b]{0.2\textwidth}
         \centering
         \includegraphics[width=1\textwidth]{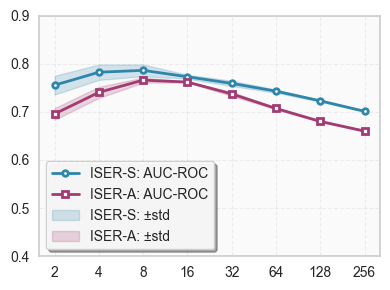}
         \caption{$\psi$}
         \label{s-psi}
     \end{subfigure}
     \begin{subfigure}[b]{0.2\textwidth}
         \centering
         \includegraphics[width=1\textwidth]{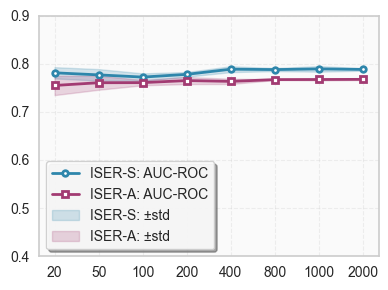}
         \caption{$t$}
         \label{s-t}
         \end{subfigure}
     \caption{Sensitivity analysis for $\psi$ and $t$ in terms of AUC-ROC.}
     \label{fig:sens}
\end{figure}

Figure~\ref{s-psi} shows that both ISER methods are sensitive to the sample size $\psi$. Performance exhibits a clear peak at moderate values of $\psi$, then gradually declines as $\psi$ increases further. When $\psi$ is too small, there are insufficient hyperspheres to effectively distinguish between normal and anomalous regions, limiting the method's discriminative ability. As $\psi$ increases beyond the optimal point, performance degrades due to excessive hypersphere partitions that create overly fine-grained divisions, leading to overfitting to the training data distribution. Figure~\ref{s-t} demonstrates that ISER methods are robust to the number of partitions $t$. Performance remains consistently stable across different values of $t$. This stability occurs because ISER uses random sampling to construct multiple isolators, and averaging across more random samples reduces variance in the final anomaly scores according to the law of large numbers. The performance quickly plateaus, indicating that moderate ensemble sizes are sufficient for stable detection while avoiding unnecessary computational overhead.

%% file: content/iseriforest.tex
\begin{figure}[!htbp]
    \centering
    \begin{subfigure}[t]{0.4\textwidth}
        \centering
        \begin{subfigure}{0.2\textwidth}
        \caption*{Data}
            \includegraphics[width=\linewidth]{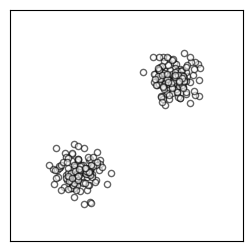}
        \end{subfigure}
        \begin{subfigure}{0.2\textwidth}
        \caption*{iForest}
            \includegraphics[width=\linewidth]{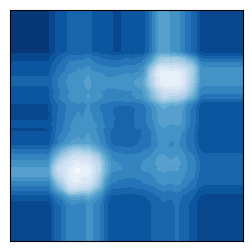}
        \end{subfigure}
        \begin{subfigure}{0.2\textwidth}
        \caption*{ISER-IF}
            \includegraphics[width=\linewidth]{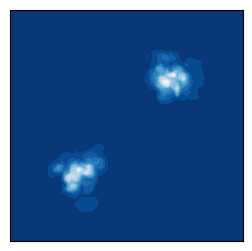}
        \end{subfigure}
        \caption{Clusters Dataset}
        \label{fig:demo-clusters}
    \end{subfigure}
    \begin{subfigure}[t]{0.4\textwidth}
        \centering
        \begin{subfigure}{0.2\textwidth}
            \includegraphics[width=\linewidth]{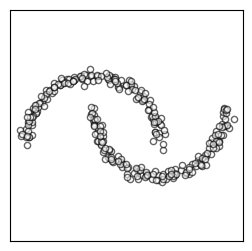}
        \end{subfigure}
        \begin{subfigure}{0.2\textwidth}
            \includegraphics[width=\linewidth]{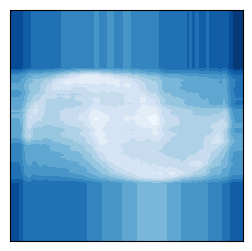}
        \end{subfigure}
        \begin{subfigure}{0.2\textwidth}
            \includegraphics[width=\linewidth]{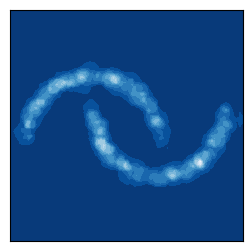}
        \end{subfigure}
        \caption{Spiral Dataset}
        \label{fig:demo-spiral}
    \end{subfigure}
    \caption{Demonstration of iForest and ISER-IF on different datasets. Left is original data, middle is standard \textbf{iForest} and right is the proposed \textbf{ISER-IF} in each figure.}\label{fig:demo}
\end{figure}

\section{ISER-based Isolation Forest}

iForest has computational efficiency but suffers from limitations that compromise its anomaly detection capability. The axis-parallel splitting criterion restricts branch cuts to hyperplanes aligned with coordinate axes, introducing systematic bias that manifests as rectangular artifacts in anomaly score maps. As demonstrated in Figure~\ref{fig:demo}, in the two-cluster scenario, standard iForest creates artificial low-score regions that extend horizontally and vertically from the cluster centers, forming `ghost' regions where points receive inappropriately low anomaly scores. The spiral dataset reveals even more dramatic limitations, where the axis-parallel cuts create vertical striping artifacts that obscure the spiral structure and fail to detect local anomalies embedded within the curved geometry.

While methods such as SCiForest~\cite{liu2010detecting} and Extended Isolation Forest~\cite{hariri2019extended} eliminate ghost regions through hyperplane cuts. To address these limitations, we propose leveraging ISER's ensemble representation as a transformed feature space for iForest. The key insight is that ISER's ensemble representation $\Phi(x)$ encodes density-aware spatial information while eliminating axis-parallel bias through its spherical partitioning scheme. By applying iForest to this transformed feature space rather than the original data space, it can inherit ISER's density awareness and local anomaly detection capability.

Once the ensemble representations are constructed using the methodology described in Section~\ref{method}, we apply iForest to the transformed feature space $\Phi(x)$. However, a critical observation emerges regarding the distribution characteristics in this transformed feature space. Anomalous points, which typically reside in sparse regions of the original space, frequently fall outside hyperspheres across multiple partitionings, resulting in ensemble representations with many components equal to 1. Even when anomalous points fall within hyperspheres, the large radii characteristic of sparse regions yield $\Phi(x)=1 - \frac{1}{r[\hat{z}_i(x)] } \approx 1$. Consequently, anomalous points are mapped to ensemble representations that exhibit high feature consistency and form tight clusters in the transformed space. In contrast, normal points residing in dense regions fall within different hyperspheres across various partitionings, and the smaller radii characteristic of dense regions produce $\Phi(x)$ that vary significantly between partitionings, resulting in ensemble representations that exhibit substantial feature diversity and appear more scattered in the transformed feature space.

\begin{figure}[!htbp]
     \centering
     \begin{subfigure}[b]{0.15\textwidth} 
         \centering
         \includegraphics[width=1\textwidth]{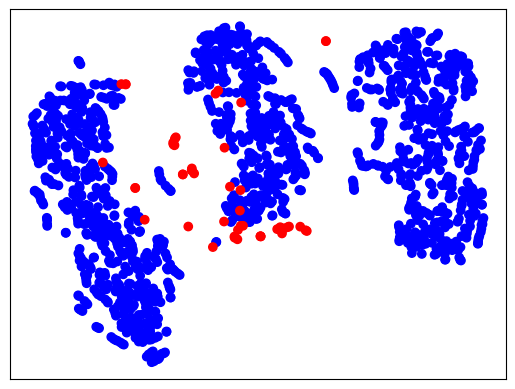} 
         \caption{Original Data.}
     \end{subfigure}
     \begin{subfigure}[b]{0.15\textwidth}
         \centering
         \includegraphics[width=1\textwidth]{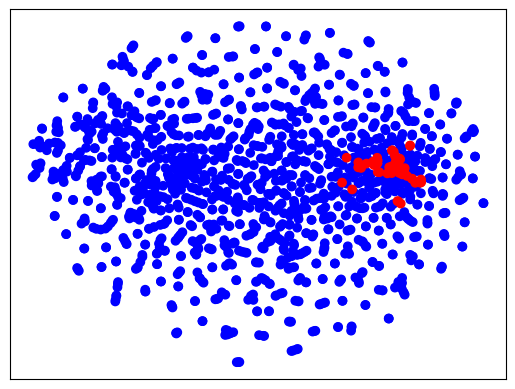}
         \caption{After Mapping.}
     \end{subfigure}
     \caption{Demonstration of ISER Mapping for the vowels dataset. In the original space (left), anomalies are scattered while normal points cluster. In $\phi$-space (right), this inverts: anomalies cluster tightly, normal points scatter.}
     \label{mapping}
\end{figure}

This creates a fundamental inversion of the iForest assumption. In the ISER feature space, normal points with diverse feature vectors are more easily separated through random cuts, leading to shorter path lengths in the isolation trees. Conversely, anomalous points with highly similar feature vectors form tight clusters that are more difficult to isolate, resulting in longer path lengths. Figure~\ref{mapping} demonstrates the feature space transformation. This modification ensures that shorter path lengths in the ISER feature space correspond to lower anomaly scores (indicating normal behavior), while longer path lengths yield higher anomaly scores (indicating anomalous behavior).

To accommodate this inversion, we modify the original iForest anomaly scoring function~\cite{liu2008isolation} as:

\begin{equation}
    S_{\text{ISER-IF}}(x, n) = 1 - 2^{-E(h(\Phi(x)))/c(n)}, \nonumber
\end{equation}
where $h(\Phi(x))$ is the path length of $\Phi(x)$ in isolation trees, $E(h(\Phi(x)))$ is the average path length, and $c(n)$ is the normalization constant for $n$ samples.

Figure~\ref{fig:demo} demonstrates the effectiveness of ISER-IF through visualization comparisons with iForest on two synthetic datasets. The results clearly show that ISER-IF produces significantly more compact and precise decision boundaries that tightly conform to the actual data distribution, completely eliminating the geometric artifacts that characterize traditional isolation methods. In the two-cluster scenario, ISER-IF completely eliminates these artifacts, producing clean, compact boundaries that precisely outline the cluster regions without any axis-aligned extensions. And in spiral dataset, ISER-IF accurately captures the spiral structure, ensuring that the central regions and gaps between spiral arms receive appropriately high anomaly scores while maintaining clean boundaries that follow the natural curvature of the data distribution. Table~\ref{tab:iser-if} demonstrates that ISER-IF achieves substantial performance improvements over iForest on datasets where iForest exhibits poor performance.



\begin{table}[!htbp]
    \centering
    \caption{Performance improvement of ISER-IF over iForest with AUC-ROC.}
    \resizebox{0.25\textwidth}{!}{
    \begin{tabular}{cllr}
    \toprule
    Data     & iForest & ISER-IF & $\uparrow$(\%)\\
    \midrule
    Ionosphere& 0.8530 & 0.9314 & 9.19\\
    glass     & 0.8016  & 0.9014 & 12.45\\
    fault     & 0.5912  & 0.7350 & 24.32 \\
    Landsat   & 0.4945  & 0.6272 & 26.84 \\
    Pima      & 0.6934  & 0.7360 & 6.14  \\
    satellite & 0.7149 & 0.7839 & 9.65   \\
    speech    & 0.4803  & 0.6443 & 34.15 \\
    vowels    & 0.7647  & 0.9526 & 24.57\\
    WPBC       &0.5215   &0.5936 & 13.83\\
    \bottomrule
    \end{tabular}}\label{tab:iser-if}
\end{table}

%% file: content/discuss.tex
\section{Discussion}

\subsection{Comparison of Scoring Methods}\label{score}
The two scoring methods differ in how they interpret the ensemble representation values. To illustrate these differences, we analyze specific examples with computed scores. 
Consider a point with ensemble values [0.9, 0.9, 0.8, 0.9, 0.1] across 5 partitions. The average-based score is 0.72, while the similarity-based score (cosine similarity with reference vector [1,1,1,1,1]) is 0.92. The average-based method is penalized by the single low value (0.1), whereas the similarity-based method emphasizes the predominant pattern of high values, interpreting this as consistent anomalous behavior with one exception due to sampling variation. For comparison, consider a point with values [0.2, 0.3, 0.3, 0.2, 0.3]. The average-based score is 0.26, appropriately reflecting intermediate anomaly levels based on moderate isolation. The similarity-based score is 0.98, as the consistent pattern closely matches the reference vector despite lower individual values. In this case, average-based scoring better captures the nuanced density-based anomaly degree, while similarity-based scoring may overestimate the anomaly level.

Although cosine similarity discards magnitude information, its strong empirical performance can be explained by the consistency of isolation patterns across partitionings. For a genuine anomaly, the value $\phi = 1$ arises structurally and repeatedly, since the
point lies off the normal manifold and is consistently left uncovered across
independent partitionings; its representation therefore points stably toward
the reference pattern. For a normal point in a dense region, the small
$\phi$ values are produced by different hyperspheres with varying radii
across partitionings and thus fluctuate, which lowers its similarity to the
reference vector. The pathological case of a perfectly uniform
intermediate-density representation, illustrated by the example above, is
rare under random sampling and is precisely the situation that average-based scoring handles well. Conversely, similarity-based scoring is robust to occasional uncovered values ($\phi = 1$) that would otherwise inflate the average. The two scores thus cover each other's failure modes, which motivates reporting both.


\subsection{Relationship and difference to Isolation-based Methods}\label{rela}


While ISER shares the similar hypersphere-based space partitioning strategy with iNNE and ADERH, the fundamental difference lies in how ensemble representations are constructed and anomaly scores are computed. 

\textbf{iNNE} computes anomaly scores using local measure rather than global measure, which is a measure of isolation of hypersphere containing $x$ relative to its neighborhood. The anomaly score is defined as $1 - r[\eta[\hat{z}_i(x)]] / r[\theta[\hat{z}_i(x)]]$, where $r[\theta[\hat{z}_i(x)]]$ is the radius of the hypersphere containing point x, and $r[\eta{[\hat{z}_i(x)}]]$ is the radius of $\theta[\hat{z}_i(x)]$'s nearest neighbor. This local measure approach suffers from a critical limitation: when the containing hypersphere and nearest hypersphere have similar radii, the method loses discriminative power regardless of the actual density characteristics. In contrast, ISER directly uses the inverse radius as a density estimation through $1 - \frac{1}{r[\hat{z}_i(x)] }$ when $x$ falls within a hypersphere, this formulation captures absolute density information rather than relative comparisons.





\textbf{ADERH} constructs hyperspheres using random pairs of points, while ISER uses the nearest neighbor distance within the sampled subset. This design choice is deliberate: the nearest neighbor radius serves as a monotonic transformation of local density, providing a theoretically grounded density encoding that directly justifies our scoring mechanism. 
In addition, ISER's per-hypersphere quantity is the radius, which is
precomputed during training, so scoring reduces to a lookup among the
covering hyperspheres. ADERH instead evaluates WPitch, a combination of
a distance ratio and a normalized density, for each covering hypersphere
at query time, and reports a consistently higher runtime than iNNE by a
constant factor. ISER achieves the same time complexity as
iNNE without this overhead.

%% file: content/conclusion.tex
\section{Conclusion}
We have proposed ISER that integrates spherical ensemble representations with density-aware scoring mechanisms. ISER uses hypersphere radii as a monotonic transformation of local density to enable effective detection of local anomalies while maintaining linear time complexity. The empirical evaluation on 20 real-world datasets demonstrates that ISER achieves superior performance compared to 12 representative baseline methods. The ability of ISER to enhance iForest through ensemble representations further validates its practical utility. An interesting future direction is to investigate whether ensemble representations can be constructed incrementally, enabling anomaly detection in streaming environments without rebuilding the hypersphere ensemble.